\documentclass{article}

\usepackage{arxiv}

\usepackage[utf8]{inputenc} 
\usepackage[T1]{fontenc}    
\usepackage{hyperref}       
\usepackage{url}            
\usepackage{booktabs}       
\usepackage{amsfonts}       
\usepackage{nicefrac}       
\usepackage{microtype}      
\usepackage{lipsum}

\usepackage[square,sort,comma,numbers]{natbib}
\usepackage{array,multirow,graphicx}
\usepackage{balance}
\usepackage[english]{babel}
\usepackage{lipsum}
\usepackage{xfrac}
\usepackage{amsmath,amssymb,amsfonts}
\usepackage{algorithmic}
\usepackage{graphicx}
\usepackage{textcomp}
\def\BibTeX{{\rm B\kern-.05em{\sc i\kern-.025em b}\kern-.08em
    T\kern-.1667em\lower.7ex\hbox{E}\kern-.125emX}}
\usepackage[utf8]{inputenc}
\usepackage{csquotes}
\usepackage{url}
\usepackage{stfloats}
\usepackage{cases}
\usepackage{array}
 \usepackage{pifont}
\newcolumntype{P}[1]{>{\centering\arraybackslash}p{#1}}
\usepackage{graphicx}

\usepackage{algorithmic}
\usepackage{subcaption}
\usepackage{caption}
\usepackage{url}

\usepackage{color}

\hypersetup{
    colorlinks=true,
    linkcolor=blue,
    filecolor=blue,      
    urlcolor=blue,
    citecolor=blue
}

\title{A Novel Multi-Step Finite-State Automaton for Arbitrarily Deterministic Tsetlin Machine Learning}

\author{
  K. Darshana Abeyrathna \\
  Centre for Artificial Intelligence Research\\
  University of Agder\\
  Grimstad, Norway \\
  \texttt{darshana.abeyrathna@uia.no} \\  
   \And
 Ole-Christoffer Granmo \\
  Centre for Artificial Intelligence Research\\
  University of Agder\\
  Grimstad, Norway \\
  \texttt{ole.granmo@uia.no} \\
   \AND
 Rishad Shafik \\
  Microsystems Research Group\\
  School of Engineering\\
  Newcastle University, UK \\
  \texttt{rishad.shafik@newcastle.ac.uk} \\
   \And 
 Alex Yakovlev \\
  Microsystems Research Group\\
  School of Engineering\\
  Newcastle University, UK \\
  \texttt{alex.yakovlev@newcastle.ac.uk} \\
   \AND 
 Adrian Wheeldon \\
  Microsystems Research Group\\
  School of Engineering\\
  Newcastle University, UK \\
  \texttt{adrian.wheeldon@newcastle.ac.uk} \\
   \And 
 Jie Lei \\
  Microsystems Research Group\\
  School of Engineering\\
  Newcastle University, UK \\
  \texttt{jie.lei@newcastle.ac.uk} \\
   \AND
 Morten Goodwin \\
  Centre for Artificial Intelligence Research\\
  University of Agder\\
  Grimstad, Norway \\
  \texttt{morten.goodwin@uia.no} \\
}

\begin{document}
\maketitle

\begin{abstract}
Due to the high energy consumption and scalability challenges of deep learning, there is a critical need to shift research focus towards dealing with energy consumption constraints. Tsetlin Machines (TMs) are a recent approach to machine learning that has demonstrated significantly reduced energy usage compared to neural networks alike, while performing competitively accuracy-wise on several benchmarks. However, TMs rely heavily on energy-costly random number generation to stochastically guide a team of Tsetlin Automata to a Nash Equilibrium of the TM game. In this paper, we propose a novel finite-state learning automaton that can replace the Tsetlin Automata in TM learning, for increased determinism. The new automaton uses multi-step deterministic state jumps to reinforce sub-patterns. Simultaneously, flipping a coin to skip every $d$'th state update ensures diversification by randomization. The $d$-parameter thus allows the degree of randomization to be finely controlled. E.g., $d=1$ makes every update random and $d=\infty$ makes the automaton completely deterministic.  Our empirical results show that, overall, only substantial degrees of determinism reduces accuracy. Energy-wise, random number generation constitutes switching energy consumption of the TM, saving up to 11 mW power for larger datasets with high $d$ values. We can thus use the new $d$-parameter to trade off accuracy against energy consumption, to facilitate low-energy machine learning.
\end{abstract}

\section{Introduction}
State-of-the-art deep learning (DL) requires massive computational resources, resulting in high energy consumption  \cite{Strubell2019EnergyAP} and scalability challenges \cite{Chen2019Edge}. There is thus a critical need to shift research focus towards dealing with energy consumption constraints \cite{GARCIAMARTIN201975}. Tsetlin Machines \cite{granmo16} (TMs) are a recent approach to machine learning (ML) that has demonstrated significantly reduced energy usage compared to neural networks alike \cite{wheeldon2020learning}. Using a linear combination of conjunctive clauses in propositional logic, the TM  has obtained competitive performance in terms of accuracy \cite{berge2019text,abeyrathna2020regression,granmo2019convolutional}, memory footprint \cite{granmo2019convolutional,wheeldon2020learning}, energy \cite{wheeldon2020learning}, and learning speed \cite{granmo2019convolutional,wheeldon2020learning} on diverse benchmarks (image classification, regression and natural language understanding). Furthermore, the 
rules that TMs build seem to be interpretable, similar to the branches in a decision tree (e.g., in the form \textbf{if} X \textbf{satisfies} condition A \textbf{and not} condition B \textbf{then} Y = 1) \cite{berge2019text}. The reported small memory footprint and low energy consumption make the TM particularly attractive for addressing the scalability and energy challenge in ML.

\textbf{Recent progress on TMs.} Recent research reports several distinct TM properties. The TM can be used in convolution, providing competitive performance on MNIST, Fashion-MNIST, and Kuzushiji-MNIST, in comparison with CNNs, K-Nearest Neighbor, Support Vector Machines, Random Forests, Gradient Boosting, BinaryConnect, Logistic Circuits and ResNet \cite{granmo2019convolutional}. The TM has also achieved promising results in text classification by using the conjunctive clauses to capture textual patterns \cite{berge2019text}. By introducing clause weights, it has been demonstrated that the number of clauses can be reduced by up to $50\times$, without loss of accuracy \cite{phoulady2020weighted}. Further, hyper-parameter search can be simplified with multi-granular clauses, eliminating the pattern specificity parameter \cite{gorji2019multigranular}. By indexing the clauses on the features that falsify them, up to an order of magnitude faster inference and learning has been reported \cite{gorji2020indexing}. Additionally, regression TMs compare favorably with Regression Trees, Random Forest Regression, and Support Vector Regression \cite{abeyrathna2020regression}. In \cite{abeyrathna2020extending}, the TM is equipped with integer weighted clauses, learnt by a stochastic searching on the line (SSL) automaton. The integer weighted TM outperforms simple Multi-Layered Artificial Neural Networks, Decision Trees, Support Vector Machines, K-Nearest Neighbor, Random Forest, Gradient Boosted Trees (XGBoost), Explainable Boosting Machines (EBMs), as well as the standard TM. 
For continuous features, a scheme for adaptive threshold-based binarization using SSLs was proposed in \cite{abeyrathna2020input}. Instead of using TAs to represent all unique thresholds, as in \cite{abeyrathna2019scheme}, merely two SSLs per feature adaptively adjust the thresholds. Finally, TMs have recently been shown to be fault-tolerant, completely masking stuck-at faults \cite{shafik2020explainability}.

\textbf{Paper Contributions.} TMs rely heavily on energy-costly random number generation to stochastically guide a team of TAs to a Nash Equilibrium of the TM game. In this paper, we propose a novel finite state learning automaton that can replace the TAs of the TM, for increased determinism. The new automaton uses multi-step deterministic state jumps to reinforce sub-patterns. Simultaneously, flipping a coin to skip every $d$'th state update ensures diversification by randomization. The $d$-parameter thus allows the degree of randomization to be finely controlled. We further evaluate the scheme empirically on five datasets, demonstrating that the new $d$-parameter can be used to trade off accuracy against energy consumption.

\section{A Multi-Step Finite-State Learning Automaton}
The origins of Learning Automata (LA) \cite{narendra3} can be traced back to the work of M. L. Tsetlin in the early 1960s \cite{Tsetlin1961}. The objective of an LA is to learn the optimal action through trial and error in a stochastic environment. Various types of LAs are available depending on the nature of the application \cite{Thathachar2004}. Due to their computational simplicity, we here focus on two-action finite-state LA, which we extend by introducing a novel periodically changing structure (variable structure).

In general, the action that an LA performs next is decided by the present state of the LA (the memory). An LA interacts with its environment iteratively. In  each  iteration, the environment randomly produces a reward or a penalty, responding to the action selected by the LA according to an unknown probability distribution. If the LA receives a reward, it reinforces the action performed by moving to a “deeper” state. If  the  action  results  in  a  penalty,  the  LA state  moves  towards  the  middle state, to weaken the performed action, ultimately jumping to the other action.  In  this  manner,  with  a  sufficient  number  of  states,  a  LA converges  to  the  action  with  the  highest  probability  of  producing rewards – the optimal action – with probability arbitrarily close to $1.0$
~\cite{narendra3}.

\begin{figure}[t]
\centering
\includegraphics[width=9cm]{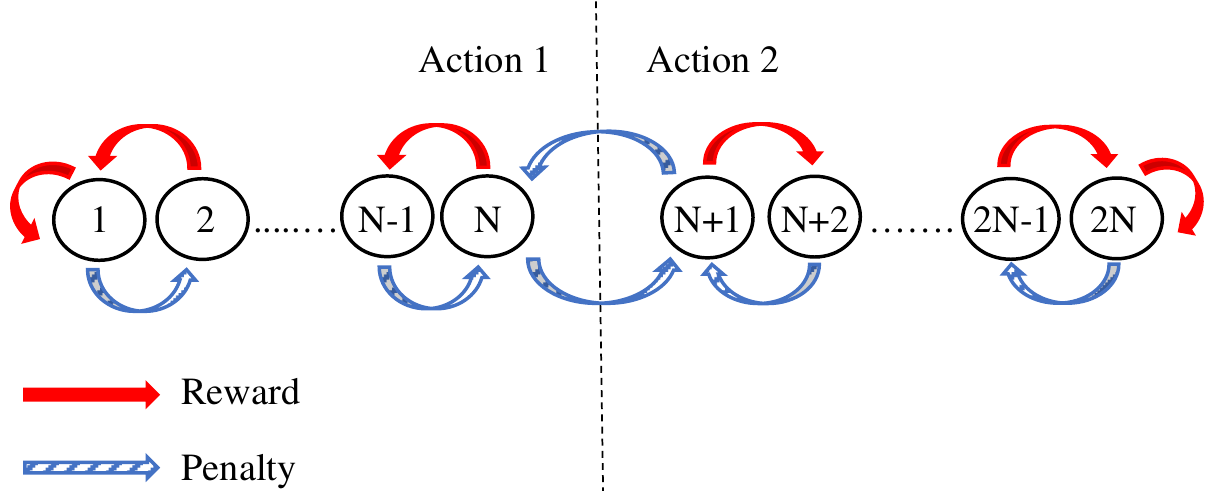}
\caption{Transition graph of a two-action Tsetlin Automaton with 2N memory states.} \label{figTA}
\end{figure}

The transitions between states can be be deterministic or stochastic. Deterministic transitions occur with probability $1.0$, while stochastic transitions are randomly performed based on a preset probability. If the transition probabilities are changing, we have a variable structure automaton, otherwise, we have one with fixed structure.  The pioneering Tsetlin Automaton (TA), depicted in Figure \ref{figTA}, is a deterministic fixed-structure finite-state automaton \cite{Tsetlin1961}. The state transition graph in the figure depicts a TA with $2N$ memory states.  States $1$ to $N$ maps to Action 1 and states $N+1$ to $2N$ maps to Action 2.

While the TA changes state in single steps, the deterministic Krinsky Automaton introduces multi-step state transitions \cite{narendra3}. The purpose is to reinforce an action more strongly when it is rewarded, and more weakly when penalized. The Krinsky Automaton behaves as a TA when the response from the environment is a penalty. However, when it is a reward, any state from $2$ to $N$ transitions to state $1$, and any state from $N+1$ to $2N-1$ transitions to state $2N$. In effect, $N$ consecutive penalties are needed to offset a single reward.

Another variant of LA is the Krylov Automaton. A Krylov Automaton makes both deterministic and stochastic single-step transitions \cite{narendra3}. The state transitions of the Krylov Automaton is identical to those of a TA for rewards. However, when it receives a penalty, it performs the corresponding TA state change randomly, with probability $0.5$.

\begin{figure}[t]
\centering
\includegraphics[width=10cm]{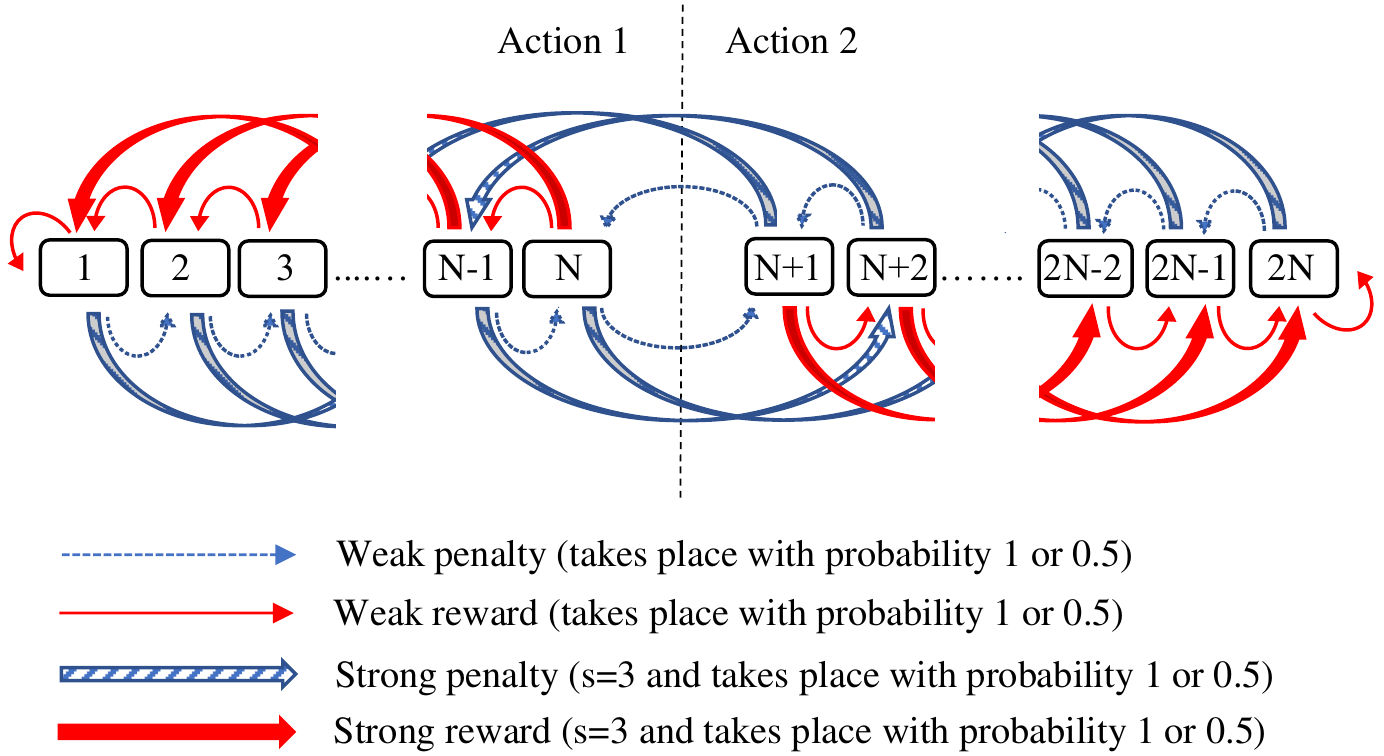}
\caption{Transition graph of the Multi-Step Variable Structure Finite-State Learning Automaton.} \label{ta}
\end{figure}

We now introduce our new type of LA, the multi-step variable-structure finite-state LA (MVF-LA), shown in Figure~\ref{ta}. The MVF-LA has two kinds of feedback, strong and weak. As covered in the next section, strong feedback is required by the TM to strongly reinforce frequent sub-patterns, while weak feedback is required to make the TM forget infrequent ones. To achieve this, weak feedback only triggers one-step transitions. Strong feedback, on the other hand, triggers $s$-step transitions. Thus, a single strong feedback is offset by $s$ instances of weak feedback. Further, MVF-LA has a variable structure that changes \emph{periodically}. That is, the MVF-LA switches between two different transition graph structures, one deterministic and one stochastic. The deterministic structure is as shown in the figure, while the stochastic structure introduces a transition probability $0.5$, for every transition. The switch between structure is performed so that every $d$'th transition is stochastic, while the remaining transitions are deterministic. 

\section{The Arbitrarily Deterministic TM (ADTM)}
\begin{figure}[t]
\centering
\includegraphics[width=10cm]{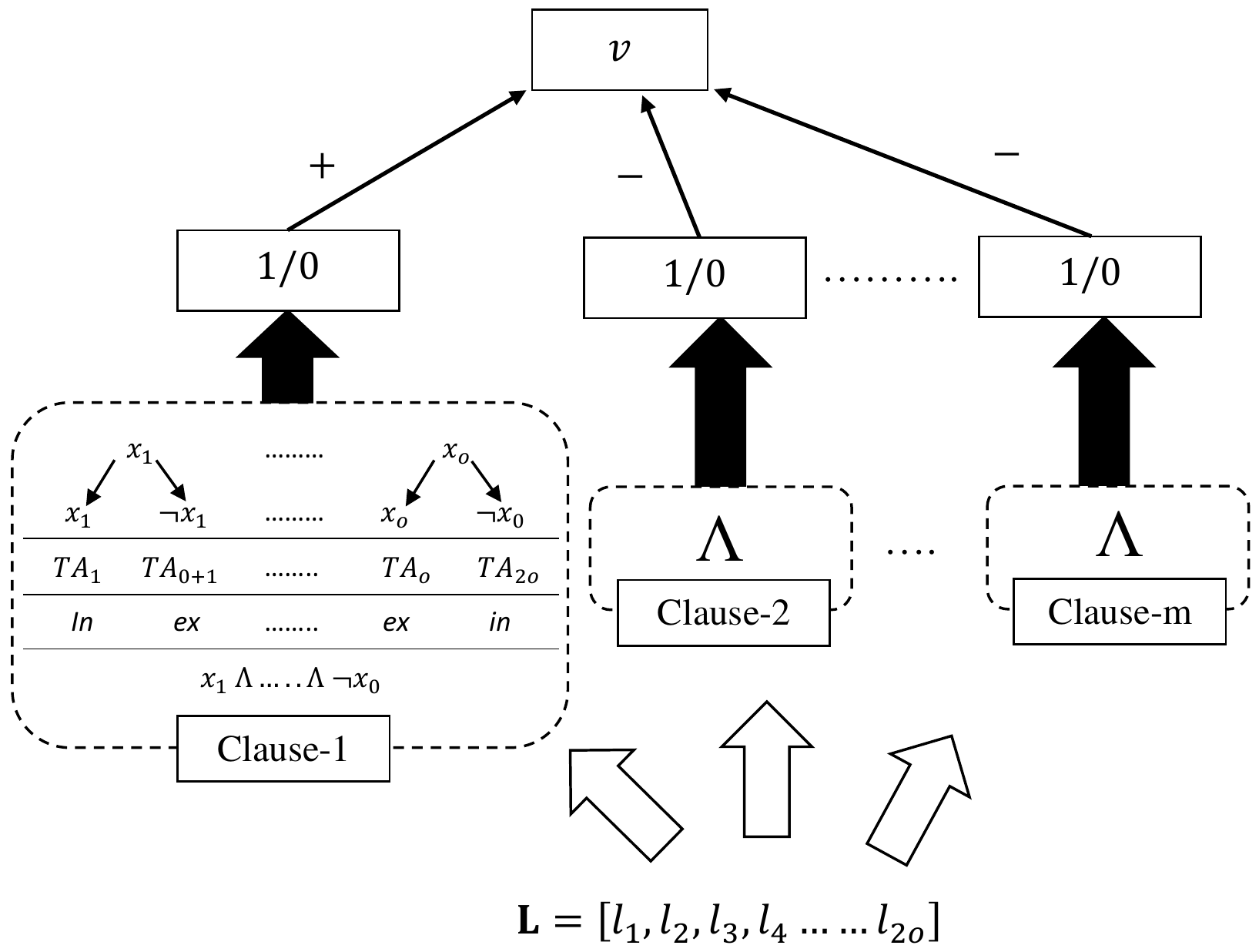}
\caption{The ADTM structure.} \label{tmstructure}
\end{figure}

In this section, we introduce the details of the ADTM, shown in Figure~\ref{tmstructure}, where the TA is replaced with the new MVF-LA. The purpose of the ADTM is to control stochasticity, thus allowing management of energy consumption.

\subsection{ADTM Inference}\label{architecture}
\textbf{Input Features.} Like the TM, an ADTM takes a feature vector of $o$ propositional variables as input, $\textbf{\textit{X}} = [x_1, x_2, x_3, \ldots, x_o]$,  to be classified into one of two classes, $y=0$ or $y=1$. These features are extended with their negation, to produce a set of literals: $\mathbf{L} = [x_1, x_2, \ldots , x_o,$  $\lnot x_1, \lnot x_2, \ldots , \lnot x_o] =$ $[l_1, l_2, \ldots, l_{2o}]$.  
 
\noindent\textbf{Clauses.} Patterns are represented by $m$ conjunctive clauses. As shown for Clause-1 in the figure, a clause in the TM comprises of $2o$ TAs, each controlling the inclusion of a specific literal. Let the set ${I}_j$, ${I}_j \subseteq \{1, \ldots, 2o\}$ denote the indexes of the literals that are included in clause $j$. When evaluating clause $j$ on input literals $\textbf{\textit{L}}$, the literals included in the clause are ANDed: $c_j = \bigwedge_{k \in {I}_j} l_k, j = 1,\dots,m$. Note that the output of an empty clause, ${I}_j = \emptyset$, is $1$ during learning and $0$ during inference.

\noindent\textbf{Classification.} In order to identify the sub-patterns associated with both of the classes of a two-class ADTM, the clauses are grouped in two. The number of clauses employed is a user set parameter $m$. Half of the clauses are assigned positive polarity ($c_j^+$). The other half is assigned negative polarity ($c_j^-$). The clause outputs, in turn, are combined into a classification decision through summation and thresholding using the unit step function $u(v) = 1 ~\mathbf{if}~ v \ge 0 ~\mathbf{else}~ 0$:
\begin{equation}
\hat{y} = u\left(\sum_{j=1}^{m/2} c_j^+(X) - \sum_{j=1}^{m/2} c_j^-(X)\right).
\end{equation}
That is, classification is based on a majority vote, with the positive clauses voting for $y=0$ and the negative for $y=1$. 

\subsection{The MVF-LA Game and Orchestration Scheme}\label{learning}
The MVF-LAs in ADTM are updated by so-called Type I and Type II feedback. Depending on the class of the current training sample $(X, y)$ and the polarity of the clause (positive or negative), the type of feedback is decided. Clauses with positive polarity receive Type I feedback when the target output is $y = 1$, and Type II feedback when the target output is $y = 0$. For clauses with negative polarity, Type I feedback replaces Type II, and vice versa. In the following, we focus only on clauses with positive polarity.

\textbf{Type I feedback:} The number of clauses which receive Type I feedback is controlled by selecting them stochastically according to Eqn. \ref{eq4}:
\begin{equation}\label{eq4}
\frac{T - \mathrm{max}(-T, \mathrm{min}(T, v))}{2T}.
\end{equation}
Above, $v = \sum_{j=1}^{m/2} c_j^+(X) - \sum_{j=1}^{m/2} c_j^-(X)$ is the aggregated clause output and $T$ is a user set parameter that decides how many clauses should be involved in learning a particular sub-pattern. 
Further, Type I feedback consists of two kinds of sub-feedback: Type~Ia and Type~Ib. Type~Ia feedback stimulates recognition of patterns by reinforcing the include action of MVF-LAs whose corresponding literal value is $1$, however, only when the clause output also is $1$. Note that an action is reinforced either by rewarding the action itself, or by penalizing the other action. Type~Ia feedback is \emph{strong}, with step size $s$ (cf. Figure~\ref{ta}). Type~Ib feedback combats over-fitting by reinforcing the \textit{exclude} actions of MVF-LAs when the corresponding literal is $0$ or when the clause output is $0$. Type~Ib feedback is \emph{weak} to facilitate learning of frequent patterns (cf. Figure~\ref{ta}).

\textbf{Type II feedback:} Clauses are also selected stochastically for receiving Type II feedback:
\begin{equation}\label{eq7}
\frac{T + \mathrm{max}(-T, \mathrm{min}(T, v))}{2T}. \\
\end{equation}
Type II feedback combats false positive clause output by seeking to alter clauses that output $1$ so that they instead output $0$. This is achieved simply by reinforcing inclusion of literals of value $0$. Thus, when the clause output is $1$ and the corresponding literal value of an MVF-LA is $0$, the include action of the MVF-LA is reinforced. Type II feedback is \emph{strong}, with step size $s$. Recall that in all of the above MVF-LA update steps, the parameter $d$ decides the determinism of the updates.

\section{Empirical Evaluation}
We now study the performance of ADTM empirically using five real-world datasets. As baselines, ADTM is compared against regular TMs and seven other state-of-the-are machine learning approaches: Artificial Neural Networks (ANNs), Support Vector Machines (SVMs), Decision Trees (DTs), K-Nearest Neighbor (KNN), Random Forest (RF), Gradient Boosted Trees (XGBoost) \cite{chen2016xgboost}, and Explainable Boosting Machines (EBMs) \cite{nori2019interpretml}. For comprehensiveness, three ANN architectures are used: ANN-1 – with one hidden layer of 5 neurons; ANN-2 – with two hidden layers of 20 and 50 neurons each, and ANN-3 – with three hidden layers and 20, 150, and 100 neurons. Performance of these predictive models are summarized in Table~\ref{othermethods}. We compute both F1-score (F1) and accuracy (Acc.) as performance measures. However, due to the class imbalance, we emphasize F1-score when comparing the performance of the different predictive models.

\subsection{Bankruptcy}
The Bankruptcy dataset contains historical records of $250$ companies\footnote{Available from \href{https://archive.ics.uci.edu/ml/datasets/qualitative\_bankruptcy}{https://archive.ics.uci.edu/ml/datasets/qualitative\_bankruptcy}.}. The outcome, Bankruptcy or Non-bankruptcy, is characterized by six categorical features. We thus binarize the features using thresholding \cite{abeyrathna2019scheme} before we feed them into the ADTM.  We first tune the hyper-parameters of the TM and the best performance is reported in Table~\ref{Bank}, for $m=100$ (number of clauses), $s=3$ (step size for MVF-LA), and $T=10$ (summation target). Each MVF-LA contains $100$ states per action. The impact of determinism is reported in Table~\ref{Bank}, for varying levels of determinism. As seen, performance is indistinguishable for  $d$-values $1$, $10$, and $100$, and the ADTM achieves its highest classification accuracy. However, notice the slight decrease of F1-score and accuracy when determinism is further increased to $500$, $1000$, and $5000$.
\begin{table}[t]
\caption{Performance of TM and ADTM with different $d$ on Bankruptcy dataset.}\label{Bank}
\centering
\newcolumntype{P}[1]{>{\centering\arraybackslash}p{#1}}
\begin{tabular}{c|P{12mm}|P{12mm}|P{12mm}|P{12mm}|P{12mm}|P{12mm}|P{12mm}}
\toprule
\multirow{2}{*}{} & \multirow{2}{*}{TM} & \multicolumn{6}{c}{ADTM}                     \\ \cline{3-8} 
                  &                     & d=1 & d=10 & d=100 & d=500 & d=1000 & d=5000 \\ \hline
F1    & 0.998 & 1.000 &	1.000 &	1.000 & 0.999 &	0.999 &	0.988 \\
Acc.    & 0.998 & 1.000 &	1.000 &	1.000 &	0.999 &	0.999 &	0.987 \\
\bottomrule
\end{tabular}
\end{table}

\begin{figure}[t]
\centering
\includegraphics[width=11cm]{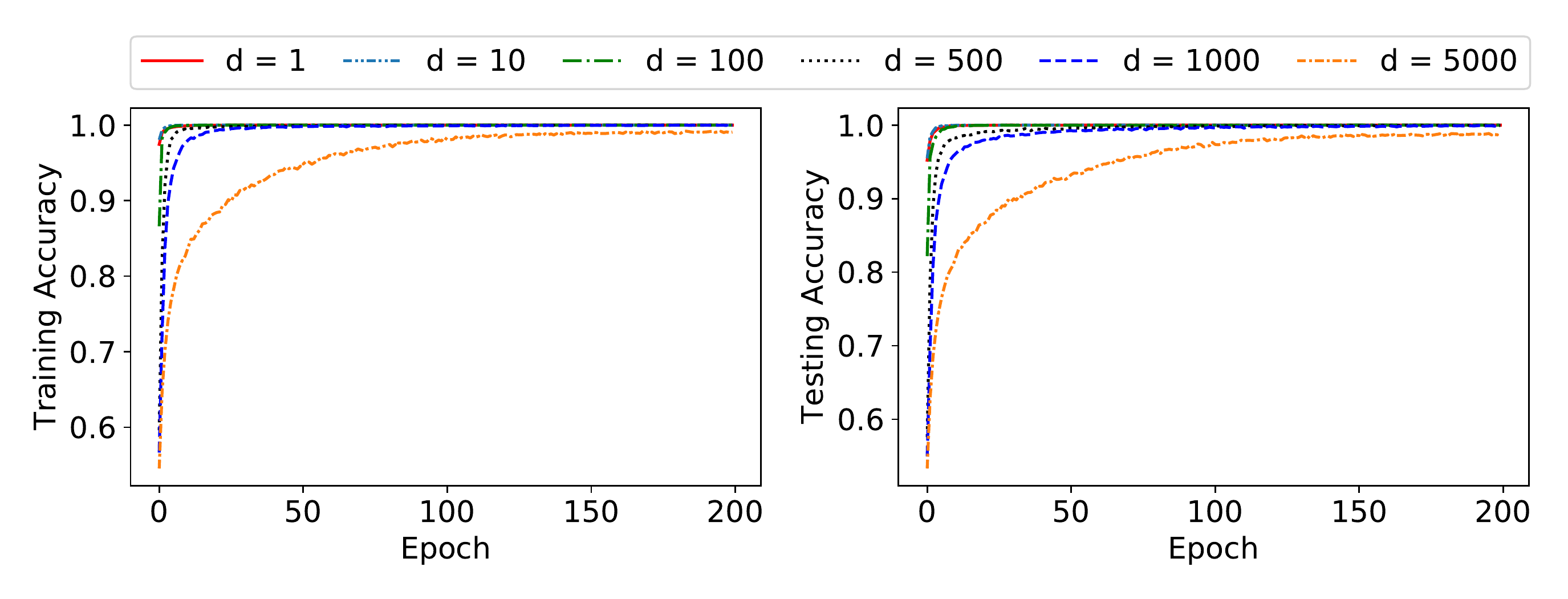}
\caption{Training and testing accuracy per epoch on the Bankruptcy dataset.} \label{banktraining}
\end{figure}

Figure~\ref{banktraining} shows how training and testing accuracy evolve over the training epochs. Only high determinism seems to influence learning speed and accuracy significantly.  The performance of the other considered machine learning models is compiled in Table \ref{othermethods}. The best performance in terms of F1-score for the other models is obtained by ANN-3. However, ANN-3 is outperformed by the ADTM for all $d$-values except when $d=5000$.

\subsection{Balance Scale}

The Balance Scale dataset\footnote{Available from \href{http://archive.ics.uci.edu/ml/datasets/balance+scale}{http://archive.ics.uci.edu/ml/datasets/balance+scale}.} contains three classes: balance scale tip to the right, tip to the left, or in balance. The class is decided by the size of the weight on both sides of the scale and the distance to each weight from the center. Hence the classes are characterized by four features. However, to make the output binary, we remove the ``balanced" class ending up with 576 data samples. The ADTM is equipped with $100$ clauses. Each MVF-LA is given $100$ states per action. The remaining two parameters, i.e., $s$ value and $T$ are fixed at $3$ and $10$, respectively. Table \ref{BS} contains the results of TM and ADTM obtained on the Balance Scale dataset. Even though ADTM uses the same number of clauses as the TM, the performance with regards to F1-score and accuracy is better with ADTM when all the MVF-LAs updates are stochastic ($d=1$). The performance of the ADTM remains the same until the determinism-parameter surpasses $100$. After that, performance degrades gradually.
\begin{table}[t]
\caption{Performance of TM and ADTM with different $d$ on Balance Scale dataset.}\label{BS}
\centering
\newcolumntype{P}[1]{>{\centering\arraybackslash}p{#1}}
\begin{tabular}{c|P{12mm}|P{12mm}|P{12mm}|P{12mm}|P{12mm}|P{12mm}|P{12mm}}
\toprule
\multirow{2}{*}{} & \multirow{2}{*}{TM} & \multicolumn{6}{c}{ADTM}                     \\ \cline{3-8} 
                  &                     & d=1 & d=10 & d=100 & d=500 & d=1000 & d=5000 \\ \hline
F1    & 0.945 & 0.982 &	0.983 &	0.982 &	0.968 &	0.951 &	0.911 \\
Acc.    & 0.948 & 0.980 &	0.981 &	0.980 &	0.935 &	0.894 &	0.793 \\
\bottomrule
\end{tabular}
\end{table}

\begin{figure}[t]
\centering
\includegraphics[width=11cm]{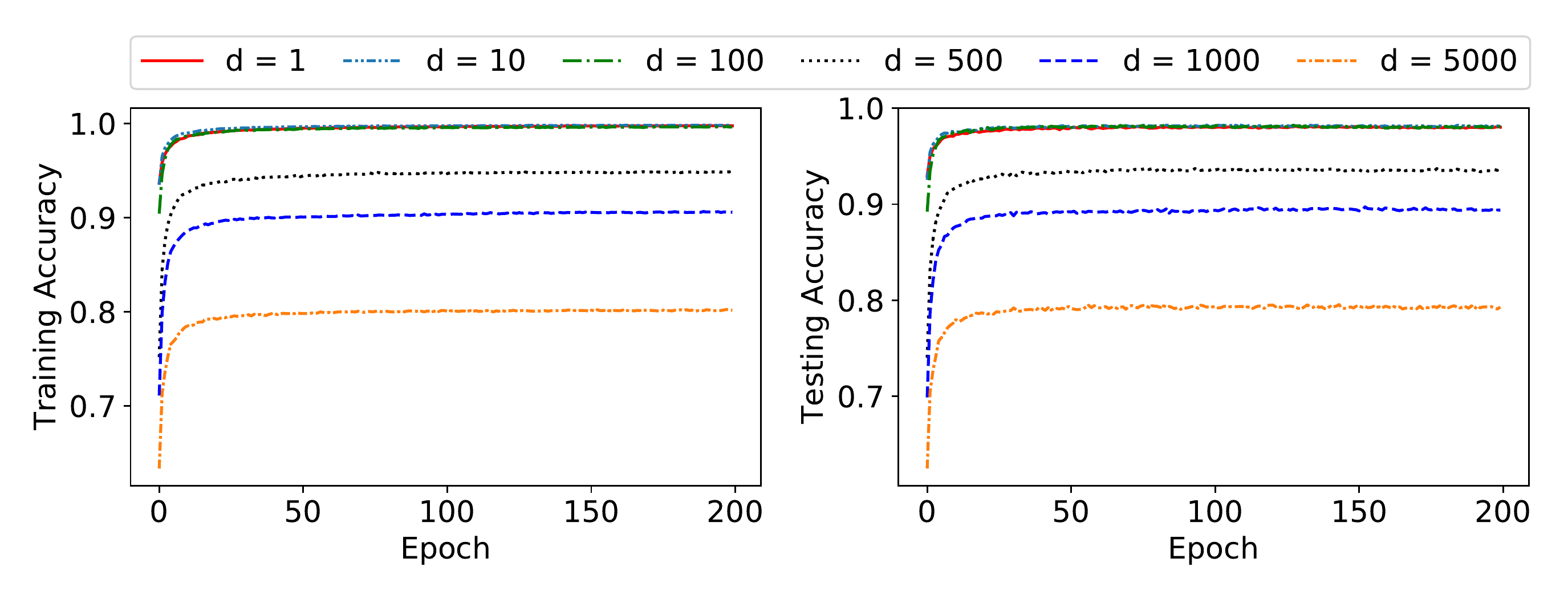}
\caption{Training and testing accuracy per epoch on the Balance Scale dataset.} \label{BStraining}
\end{figure}

Progress of training and testing accuracy per epoch can be found in Figure~\ref{BStraining}. Each ADTM setup reaches its peak training and testing accuracy and becomes stable within a fewer number of training epochs. As can be seen, accuracy is maintained up to $d=100$, thus reducing random number generation to $1\%$ without accuracy loss.
From the results listed in Table \ref{othermethods} for the other machine learning approaches, EBM achieves the highest F1-score and accuracy.

\subsection{Breast Cancer}
The Breast Cancer dataset\footnote{Available from \href{https://archive.ics.uci.edu/ml/datasets/Breast+Cancer}{https://archive.ics.uci.edu/ml/datasets/Breast+Cancer}} contains 286 patients records related to the recurrence of breast cancer (201 with non-recurrence and 85 with recurrence). The recurrence of breast cancer has to be estimated using nine features: Age, Menopause, Tumor Size, Inv Nodes, Node Caps, Deg Malig, Side (left or right), the Position of the Breast, and whether Irradiated or not. However, some of the patient samples miss some of the feature values. These samples are removed from the dataset in the present experiment.

The ADTM is arranged with the following parameter setup: $m=100$, $s=5$, $T=10$, and the number of states in MVF-LA per action is $100$. 
The classification accuracy of the TM and ADTM are summarized in Table \ref{BC}. In contrast to the previous two datasets, the performance of both TM and ADTM is considerably lower, and further decreases with increasing determinism. However, the F1 measures obtained by all the other considered machine learning models are also low, i.e., less than $0.500$. The highest F1-score is obtained by ANN-1 and KNN, which both are lower than what is achieved with ADTM for $d$-values up to $100$.
\begin{table}[t]
\caption{Performance of TM and ADTM with different $d$ on Breast Cancer dataset.}\label{BC}
\centering
\newcolumntype{P}[1]{>{\centering\arraybackslash}p{#1}}
\begin{tabular}{c|P{12mm}|P{12mm}|P{12mm}|P{12mm}|P{12mm}|P{12mm}|P{12mm}}
\toprule
\multirow{2}{*}{} & \multirow{2}{*}{TM} & \multicolumn{6}{c}{ADTM}                     \\ \cline{3-8} 
                  &                     & d=1 & d=10 & d=100 & d=500 & d=1000 & d=5000 \\ \hline
F1    & 0.531 & 0.568 &	0.531 &	0.501 &	0.490 &	0.501 &	0.488 \\
Acc.    & 0.703 & 0.702 &	0.698 &	0.691 &	0.690 &	0.690 &	0.693 \\
\bottomrule
\end{tabular}
\end{table}

\begin{figure}[t] 
\centering
\includegraphics[width=11cm]{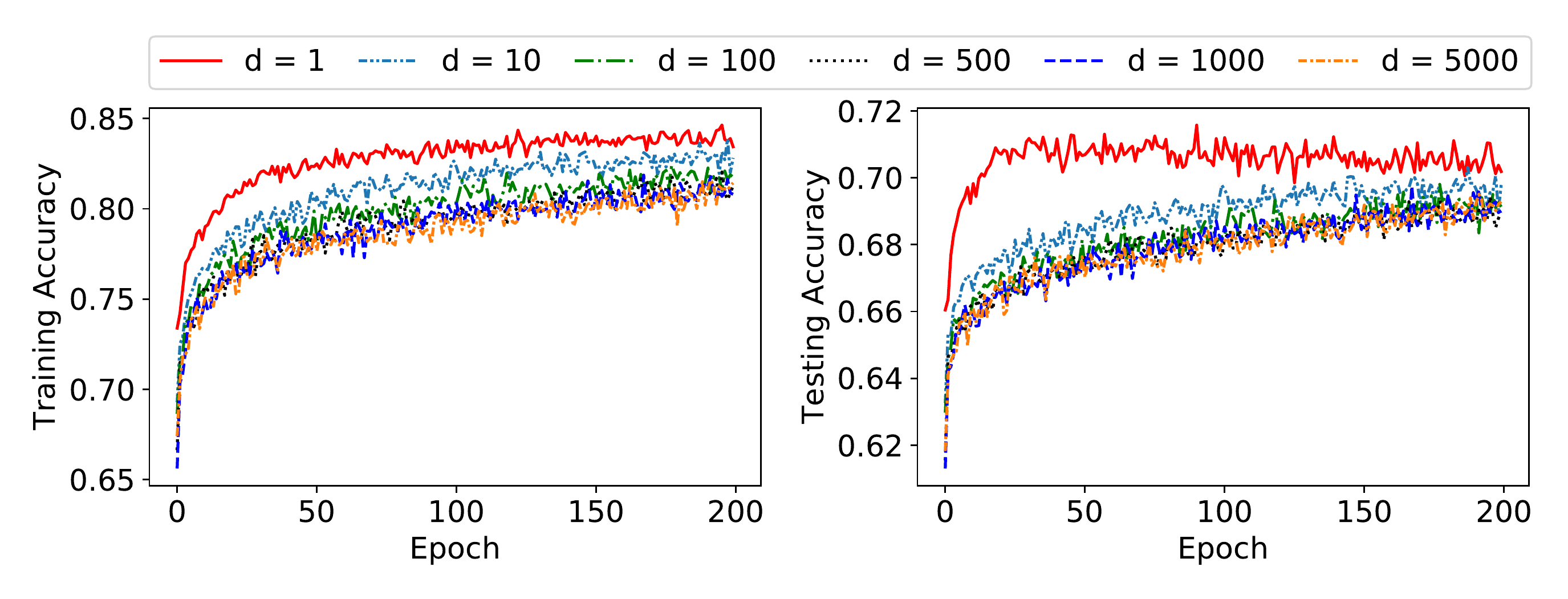}
\caption{Training and testing accuracy per epoch on the Breast Cancer dataset.} \label{BCtraining}
\end{figure}

The training and testing accuracy progress per epoch is reported in Figure~\ref{BCtraining}, showing a clear degradation of performance with increasing determinism.

\begin{table}[!ht]
\caption{Performance of TM and ADTM with different $d$ on Liver Disorders dataset.}\label{LD}
\centering
\newcolumntype{P}[1]{>{\centering\arraybackslash}p{#1}}
\begin{tabular}{c|P{12mm}|P{12mm}|P{12mm}|P{12mm}|P{12mm}|P{12mm}|P{12mm}}
\toprule
\multirow{2}{*}{} & \multirow{2}{*}{TM} & \multicolumn{6}{c}{ADTM}                     \\ \cline{3-8} 
                  &                     & d=1 & d=10 & d=100 & d=500 & d=1000 & d=5000 \\ \hline
F1    & 0.648 & 0.705 &	0.694 &	0.692 &	0.692 &	0.689 &	0.692 \\
Acc.    & 0.533 & 0.610 &	0.610 &	0.612 &	0.612 &	0.610 &	0.611 \\
\bottomrule
\end{tabular}
\end{table}

\begin{figure}[!ht]
\centering
\includegraphics[width=11cm]{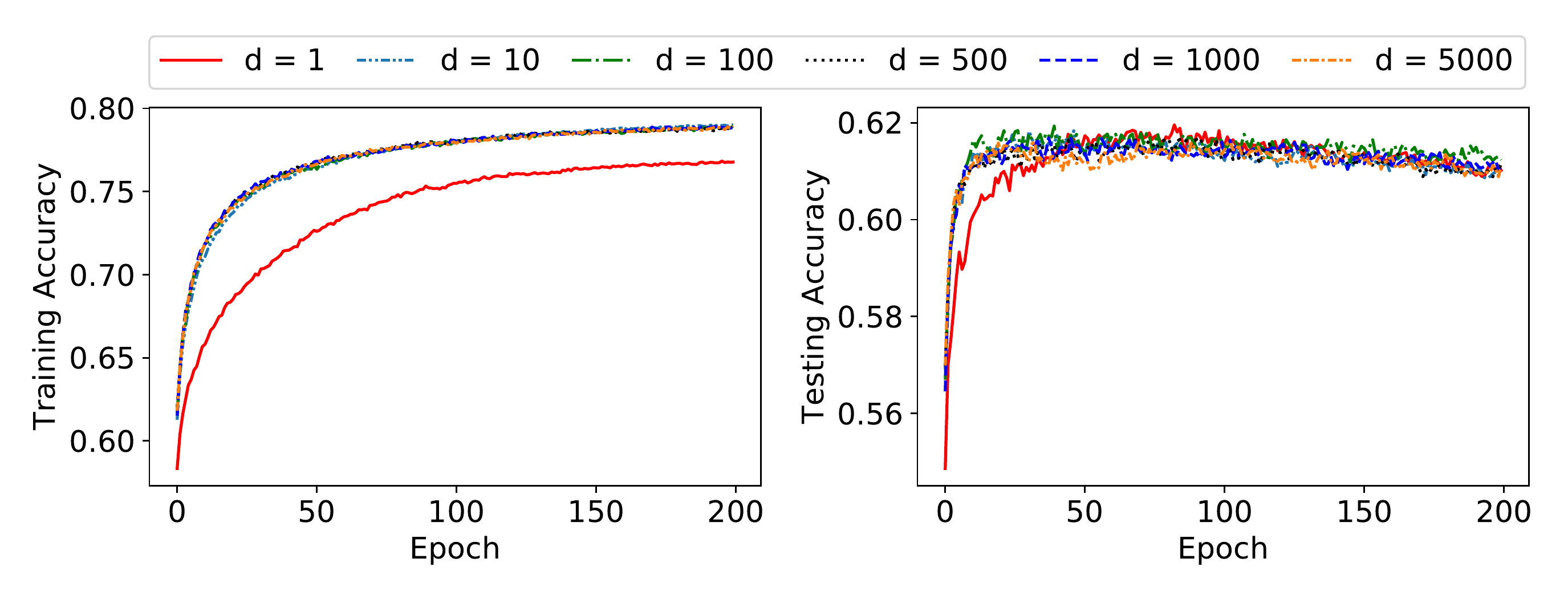}
\caption{Training and testing accuracy per epoch on the Liver Disorders dataset.} \label{LDtraining}
\end{figure}

\subsection{Liver Disorders}

The Liver Disorders dataset\footnote{Available from \href{https://archive.ics.uci.edu/ml/datasets/Liver+Disorders}{https://archive.ics.uci.edu/ml/datasets/Liver+Disorders}.} was created by BUPA Medical Research and Development Ltd. (hereafter “BMRDL”) during the 1980s as part of a larger health-screening database. The dataset consists of 7 attributes. However, McDermott and Forsyth \cite{mcdermott2016diagnosing} claim that many researchers have used the dataset incorrectly, considering the Selector attribute as the class label. Based on the recommendation of McDermott and Forsythof, we here instead use the Number of Half-Pint Equivalents of Alcoholic Beverages as the dependent variable, binarized using the threshold $\geq 3$. The Selector attribute is discarded. The remaining attributes represent the results of various blood tests, and we use them as features. 

Here, ADTM is given $10$ clauses per class, with $s=3$ and $T=10$. Each MVF-LA action possesses $100$ states.  The performance of ADTM for different levels of determinism is summarized in Table \ref{LD}. For $d=1$, the F1-score of ADTM is better than what is achieved with the standard TM. In contrast to the performance on previous datasets, the performance of ADTM on Liver Disorders dataset with respect to F1-score does not decrease significantly with $d$. Instead, it fluctuates around $0.690$. Unlike the other datasets, the ADTM with $d=1$ actually requires more training rounds than for larger $d$-values, before it learns the final MVF-LA actions. The ADTM with $d=1$ is also unable to reach a similar level of training accuracy, compared to higher $d$-values. Despite the diverse learning speed, testing accuracy becomes similar after roughly $50$ training rounds.
The other considered machine learning models obtain somewhat similar F1-scores on the same dataset, with DT, RF, and EBM peaking with scores higher than $0.700$. 

\subsection{Heart Disease}
The Heart Disease dataset\footnote{Available from \href{https://archive.ics.uci.edu/ml/datasets/Statlog+\%28Heart\%29}{https://archive.ics.uci.edu/ml/datasets/Statlog+\%28Heart\%29}.} concerns prediction of heart disease. To this end,  13 features are available, selected among 75. Out of the 13 features, 6 are real-valued, 3 are binary, 3 are nominal, and one is ordered.
In this case, the ADTM is built on $100$ clauses. The number of state transitions when the feedback is strong, $s$, is equal to $3$ while the target, $T$, is equal to $10$. The number of states per MVF-LA action in the ADTM is $100$. 
As one can see in Table \ref{HD}, the ADTM provides better performance than TM in terms of F1-score and accuracy when $d=1$. The F1-measure increases with the $d$-value and peaks at $d=100$. Then it fluctuates and reaches its lowest value of $0.605$ when $d=5000$.
\begin{table}[t]
\caption{Performance of TM and ADTM with different $d$ on Heart Disease dataset. }\label{HD}
\centering
\newcolumntype{P}[1]{>{\centering\arraybackslash}p{#1}}
\begin{tabular}{c|P{12mm}|P{12mm}|P{12mm}|P{12mm}|P{12mm}|P{12mm}|P{12mm}}
\toprule
\multirow{2}{*}{} & \multirow{2}{*}{TM} & \multicolumn{6}{c}{ADTM}                     \\ \cline{3-8} 
                  &                     & d=1 & d=10 & d=100 & d=500 & d=1000 & d=5000 \\ \hline
F1    & 0.687 & 0.759 &	0.766 &	0.767 &	0.760 &	0.762 &	0.605 \\
Acc.    & 0.672 & 0.778 &	0.780 &	0.783 &	0.773 &	0.781 &	0.633 \\
\bottomrule
\end{tabular}
\end{table}

\begin{figure}[t]
\centering
\includegraphics[width=11cm]{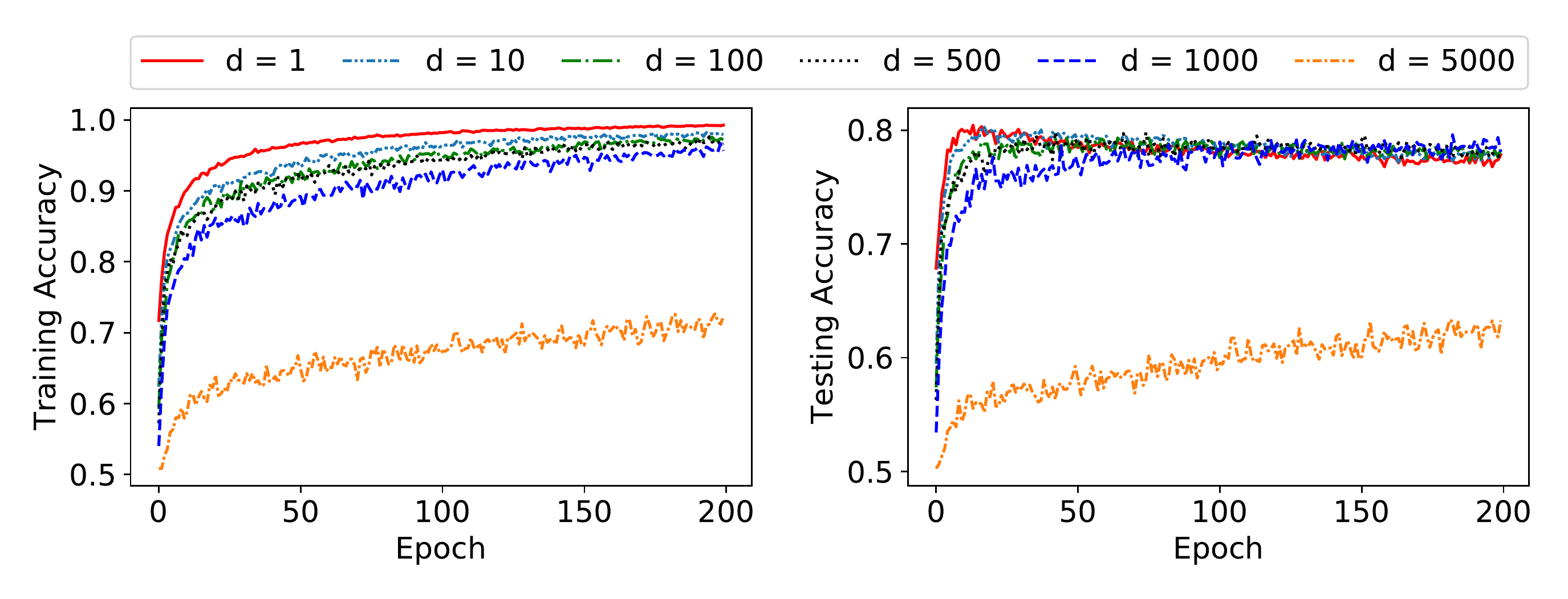}
\caption{Training and testing accuracy per epoch on the Heart Disease dataset.} \label{HDtraining}
\end{figure}

Figure~\ref{HDtraining} shows that the training and testing accuracy for $d=5000$ is consistently low over the training epochs, compared to the training and testing accuracy of other $d$-values. The other $d$-values, after initial variations, perform similarly roughly from the $75^{th}$ epoch and onward.
Out of other machine learning algorithms, EBM provides the best F1-score. Even though ANN-1, ANN-2, DT, RF, and XGBoost obtain better F1-scores than TM, the F1 scores of ADTM when $d$ equals to $1$, $10$, $100$, $500$, and $1000$ are the highest ones.

\begin{table}[ht]
\caption{Classification accuracy of state-of-the-art machine learning models on all the datasets.}\label{othermethods}
\centering
\newcolumntype{P}[1]{>{\centering\arraybackslash}p{#1}}
\begin{tabular}{c|P{9.5mm}|P{9.5mm}|P{9.5mm}|P{9.5mm}|P{9.5mm}|P{9.5mm}|P{9.5mm}|P{9.5mm}|P{9.5mm}|P{9.5mm}}
\toprule
\multirow{2}{*}{} & \multicolumn{2}{c|}{Bankruptcy} & \multicolumn{2}{c|}{Balance Scale} & \multicolumn{2}{c|}{Breast Cancer} & \multicolumn{2}{c|}{Liver Disorder} & \multicolumn{2}{c}{Heart Disease} \\ \cline{2-11} 
                  & F1             & Acc.           & F1               & Acc.            & F1               & Acc.            & F1                & Acc.             & F1               & Acc.            \\ \hline
ANN-1             & 0.995          & 0.994          & 0.990            & 0.990           & 0.458            & 0.719           & 0.671             & 0.612            & 0.738            & 0.772           \\ 
ANN-2             & 0.996          & 0.995          & 0.995            & 0.995           & 0.403            & 0.683           & 0.652             & 0.594            & 0.742            & 0.769           \\ 
ANN-3             & 0.997          & 0.997          & 0.995            & 0.995           & 0.422            & 0.685           & 0.656             & 0.602            & 0.650            & 0.734           \\ 
DT                & 0.993          & 0.993          & 0.986            & 0.986           & 0.276            & 0.706           & 0.728             & 0.596            & 0.729            & 0.781           \\ 
SVM               & 0.994          & 0.994          & 0.887            & 0.887           & 0.384            & 0.678           & 0.622             & 0.571            & 0.679            & 0.710           \\ 
KNN               & 0.995          & 0.994          & 0.953            & 0.953           & 0.458            & 0.755           & 0.638             & 0.566            & 0.641            & 0.714           \\ 
RF                & 0.949          & 0.942          & 0.859            & 0.860           & 0.370            & 0.747           & 0.729             & 0.607            & 0.713            & 0.774           \\ 
XGBoost           & 0.983          & 0.983          & 0.931            & 0.931           & 0.367            & 0.719           & 0.656             & 0.635            & 0.701            & 0.788           \\ 
EBM               & 0.993          & 0.992          & 1.000            & 1.000           & 0.389            & 0.745           & 0.710             & 0.629            & 0.783            & 0.824           \\ 
\bottomrule
\end{tabular}
\end{table}

\begin{table}[!ht]
    \centering
    \caption{Comparative power per datapoint with two different $d$ values.}
    \begin{tabular}{l|c|c|c|c|c}
    \toprule
      & Bankruptcy & Breast Cancer & Balance Scale & Liver Disorder & Heart Disease\\ \hline
     Power (d=1) & 6.94 \textit{mW} & 15.8 \textit{mW} & 7.7 \textit{mW} & 12.6 \textit{mW} & 148 \textit{mW}\\ 
     Power (d=5000) & 6.45 \textit{mW} & 14.7 \textit{mW} & 7.2 \textit{mW} & 11.8 \textit{mW} & 137.6 \textit{mW}\\ 
    \bottomrule
    \end{tabular}
    \label{tab:power}
\end{table}

\section{Effects of Determinism on Energy Consumption}

In the hardware implementation of TM, power is consumed by the pseudorandom number generators (PRNGs) when generating a new random number~\cite{wheeldon2020learning}.
This is referred to as \emph{switching power}.
In the TM, every TA update is randomized, and switching power is consumed by the PRNGs on every cycle.
Additionally, power is also consumed by the PRNGs whilst idle. We term this \emph{leakage power}.
Leakage power is always consumed by the PRNGs whilst they are powered up, even when not generating new numbers.

In the ADTM with hybrid TA where the determinism parameter $d$ is introduced, $d=1$ would be equivalent to a TM where every TA update is randomized.
$d=\infty$ means the ADTM is fully deterministic, and no random numbers are required from the PRNG.
If a TA update is randomized only on the $d^\textrm{th}$ cycle, the PRNGs need only be actively switched (and therefore consume \emph{switching power}) for $\frac{1}{d}$ portion of the entire training procedure.
The switching power consumed by the PRNGs accounts for 7\% of the total system power when using a traditional TA (equivalent to $d=1$). With $d=100$ this is reduced to 0.07\% of the system power, and with $d=5000$ this is reduced further to 0.001\% of the same. It can be seen that as $d$ increases in the ADTM, the switching power consumed by the PRNGs tends to zero.

In the special case of $d=\infty$ the PRNGs are no longer required for TA updates since the TAs are fully deterministic -- we can omit these PRNGs from the design and prevent their \emph{leakage power} from being consumed.
The leakage power of the PRNGs accounts for 32\% of the total system power. On top of the switching power savings this equates to 39\% of system power, meaning large power and therefore energy savings can be made in the ADTM.

Table~\ref{tab:power} shows comparative training power consumption per datapoint (i.e. all TAs being updated concurrently) for two different $d$ values: $d$=1 and $d$=5000. Typically, the overall power is higher for bigger datasets as they require increased number of concurrent TAs as well as PRNGs. As can be seen, the increase in $d$ value reduces the power consumption by 11 mW in the case of Heart Disease dataset. This saving is made by reducing the switching activity in the PRNGs as explained above. More savings are made by larger $d$ values as the PRNG concurrent switching activities are reduced.

\section{Conclusion}

In this paper, we proposed a novel finite-state learning automaton (MFV-LA) that can replace the Tsetlin Automaton in TM learning, for increased determinism, and thus reduced energy usage. The new automaton uses multi-step deterministic state jumps to reinforce sub-patterns. Simultaneously, flipping a coin to skip every $d$'th state update ensures diversification by randomization. The new $d$-parameter thus allows the degree of randomization to be finely controlled. E.g., $d=1$ makes every update random and $d=\infty$ makes the automaton fully deterministic. Our empirical results show that, overall, only substantial degrees of determinism reduces accuracy. Energy-wise, the pseudorandom number generator contributes to switching energy consumption within the TM, which can be completely eliminated with $d=\infty$. We can thus use the new $d$-parameter to trade off accuracy against energy consumption, to facilitate low-energy machine learning.

\section*{Acknowledgement}
The authors gratefully acknowledge the contributions from Jonathan Edwards at Temporal Computing on strategies for deterministic Tsetlin Machine learning.

\bibliographystyle{unsrt}  
\bibliography{sample}  
\end{document}